\documentclass{article}

\usepackage{arxiv}

\usepackage[utf8]{inputenc} 
\usepackage[T1]{fontenc}    
\usepackage{hyperref}       
\usepackage{url}            
\usepackage{booktabs}       
\usepackage{amsfonts}       
\usepackage{nicefrac}       
\usepackage{microtype}      
\usepackage{lipsum}
\usepackage{graphicx}
\graphicspath{ {./images/} }
\usepackage{times}
\usepackage{latexsym}

\usepackage[T1]{fontenc}
\usepackage[utf8]{inputenc}

\usepackage{microtype}

\usepackage{graphicx}
\usepackage{amsmath,amssymb,amsfonts}
\usepackage{algorithmic}
\usepackage{graphicx}
\usepackage{textcomp}
\usepackage{xcolor}
\usepackage{times}
\usepackage{latexsym}
\usepackage{booktabs}
\usepackage{subcaption}
\usepackage{array}
\usepackage{float}
\usepackage{hyperref}
\usepackage[htt]{hyphenat}
\usepackage{amsmath}
\usepackage{multirow}

\DeclareMathOperator*{\argmax}{arg\,max}

\title{MM-PoE: Multiple Choice Reasoning via. Process of Elimination using Multi-Modal Models}

\author{
 Sayak Chakrabarty$^*$ \\
  Department of Computer Science and Engineering\\
  Northwestern University\\
  Evanston, IL 60208 \\
  \texttt{sayakchakrabarty2025@u.northwestern.edu} \\
   \And
 Souradip Pal$^*$ \\
  School of Electrical \& Computer Engineering\\
  Purdue University\\
  West Lafayette, IN 47906 \\
  \texttt{pal43@purdue.edu} \\
  }

\begin{document}
\def\thefootnote{*}\footnotetext{These authors contributed equally to this work}\def\thefootnote{\arabic{footnote}}
\maketitle

\begin{abstract}
This paper introduces Multiple Choice Reasoning via. Process of Elimination using Multi-Modal models, herein referred to as \textit{Multi-Modal Process of Elimination (MM-PoE)}. This novel methodology is engineered to augment the efficacy of Vision-Language Models (VLMs) in multiple-choice visual reasoning tasks. Diverging from conventional approaches that evaluate each option independently, MM-PoE employs a dual-step scoring paradigm that initially identifies and excludes implausible choices, subsequently concentrating on the most probable remaining options. This method emulates human test-taking strategies, where individuals typically eliminate clearly incorrect answers prior to selecting the optimal response. Our empirical evaluations, conducted across three benchmark datasets, reveal that MM-PoE significantly improves both zero-shot and few-shot performance of contemporary state-of-the-art VLMs. Critically, this approach not only broadens the application of the elimination process to multi-modal contexts but also allows few-shot experiments, thereby addressing two principal limitations concerning usage of PoE only in zero-shot settings and only with a language-only framework. As a result, MM-PoE not only refines the reasoning capabilities of VLMs but also broadens their applicability to complex visual question-answering scenarios. All code and documentation supporting our work are available at \url{https://pypi.org/project/mm-poe/}, enabling researchers and practitioners to easily integrate and further develop these techniques.
\end{abstract}

\keywords{Generative AI \and Natural Language Processing\and Large Language Models\and Vision Language Models\and Multiple Choice Reasoning \and Visual Question Answering \and Few-shot Prompting}

\section{Introduction}
Large Language Models (LLMs) have shown remarkable in-context learning capabilities \cite{chakrabarty2024free,malkin2022coherence,ouyang2022training,shum2023automatic,srivastava2023beyond,suzgun2023challenging,touvron2023llama,min2022noisy,kojima2022large,lyu2023z,wei2022chain,ye2023satisfiability,zhao2021calibrate,lu2022learn}, including proficiency in multiple-choice reasoning tasks \cite{ma2023poe,robinson2023leveraging,wang2023bself}. However, while humans often engage in a process of elimination discarding obviously incorrect choices before selecting the final answer\cite{sap2019social,du2020event,freivalds2002learning}, current models do not inherently emulate this approach. The same limitation applies to Vision-Language Models (VLMs), especially in scenarios where a question-answering system must consider both textual and visual information simultaneously. This discrepancy can limit the effectiveness of vision language models in accurately solving such tasks. 

To address this gap, we present the \textit{Multi-Modal Process of Elimination (MM-PoE)}, a two-step scoring method that closely mimics human reasoning strategies in multi-modal settings. First, MM-PoE scores and filters out likely incorrect answers by masking. Next, it re-scores the remaining subset of options to identify the most plausible one. By incorporating a structured elimination step, MM-PoE improves the models' ability to handle complex visual reasoning tasks and yields more accurate, interpretable results. Our zero-shot and few-shot experiments\cite{freivalds2002learning,nie2020adversarial,wang2023bself,kojima2022large} demonstrate the effectiveness of MM-PoE on three datasets encompassing a range of visual reasoning challenges\cite{talmor2019commonsenseqa,seo2022debiasing,Kembhavi2016ADI}. The results show that MM-PoE outperforms conventional baselines and is straightforward to integrate with contemporary state-of-the-art VLM architectures. This not only improves performance but also helps in making the decision-making process more transparent and aligned with human cognitive strategies. Using this tool, researchers and practitioners can experiment and significantly improve the accuracy and reliability of VLMs in multiple choice reasoning tasks, making it valuable for advancing machine learning models for visual reasoning\cite{Kembhavi2016ADI,du2020event,robinson2023leveraging,seo2022debiasing}.

\subsection{Background and Related Work}
\vspace{-1mm}
The rapid advancement in machine learning\cite{chakrabarty2023single,zhang2022new,zhang2023dynamically} techniques have been fueled by training large models\cite{yao2023tree,fei2023mitigating,kojima2022large,ouyang2022training,suzgun2023challenging,zhao2021calibrate} on massive multi-modal datasets like \cite{VQA}. In the context of multiple-choice question-answering, humans commonly employ a process of elimination, systematically removing unlikely answers to streamline their decision\cite{sap2019social,talmor2019commonsenseqa,freivalds2002learning}. Such reasoning patterns have been explored in other domains, for example, through logical inference methods \cite{Kembhavi2016ADI,bolonkin2024judicial,wang2023pinto,sap2019social}, and have demonstrated that aligning computational approaches with human strategies can be beneficial.

Despite their impressive capabilities, both LLMs and VLMs like \cite{GIT_Base_VQA,GIT_Base_TextVQA,PaliGemma_AI2D,PaliGemma_ScienceQA,PaliGemma_VQAV2,Idefics2} typically consider each candidate solution independently or rely on aggregated likelihood without explicitly discarding less plausible options\cite{zhao2021calibrate,ye2023satisfiability,yao2023tree,wei2022chain,du2020event,kojima2022large}. This can be problematic in visually complex tasks, where multiple distractors may appear superficially similar. Prior research \cite{holtzman2021surface,sap2019social} has indicated that more human-like reasoning frameworks can enhance performance, suggesting that incorporating elimination steps could address known limitations \cite{ma2023poe}. Our hypothesis posits that vision language models, when equipped with a mechanism to discard implausible answers systematically, can achieve better performance on multiple-choice visual reasoning tasks\cite{robinson2023leveraging,lyu2023z}. This is particularly relevant in the context of logical reasoning, where the elimination of clearly incorrect options can simplify the decision process and potentially lead to more accurate outcomes.

Our MM-PoE approach grounded in everyday test-taking strategies leverages these insights, focusing on structured elimination prior to final selection. The recent success of large-scale vision-language models \cite{brown2020language} and foundational architectures for multi-modal reasoning, such as BLIP-2 \cite{BLIP2_Opt}, \cite{BLIP2_FLAN}, provides a strong basis to improve upon. By integrating a two-step reasoning mechanism, we align the decision-making process with human heuristics, ultimately leading to more accurate and interpretable outcomes. MM-PoE aims to refine the ability to eliminate unlikely choices, making it more targeted and efficient in dealing with the nuanced challenges presented by multiple-choice questions, thus showing that by mimicking human reasoning processes, we can make VLMs not only perform better on standardized visual reasoning tasks but also behave in ways that are more interpretable and aligned with human cognitive processes. This approach also resolves one of the key limitations mentioned in \cite{ma2023poe} and thus extends their language-based model to a multimodal(language and vision) framework.

\subsection{Our Contributions}

\begin{itemize}\setlength{\itemsep}{0pt} 
\item This study demonstrates that the Multi-Modal Process of Elimination (MM-PoE) enhances performance across three benchmark datasets, with notable efficacy in logical and visually grounded multiple-choice question-answering tasks. 
\item It also establishes that MM-PoE can be seamlessly integrated with existing state-of-the-art Vision Language Models (VLMs), extending the conceptual framework of \cite{ma2023poe} to a more comprehensive multi-modal context. This addresses the previous limitation of focusing on a single modality as noted in \cite{ma2023poe}. 
\item While the preceding research in \cite{ma2023poe} was confined to zero-shot\cite{kojima2022large} settings and did not address few-shot scenarios, this work extends the application to few-shot settings as well, thereby broadening the scope of the original findings. 
\end{itemize}

\section{Methodology}
The Multi-Modal Process of Elimination (MM-PoE) introduced in this paper operates on a two-step mechanism designed to enhance the decision-making capabilities of vision language models (VLMs) in multiple-choice visual reasoning tasks. This method employs a novel approach to option elimination followed by a focused prediction phase. The strategy is rooted in the belief that separating the elimination of clearly incorrect options from the choice of the best remaining option will improve overall task performance.

Given a multiple-choice visual reasoning task, we define the problem setting as follows:

\begin{itemize}\setlength{\itemsep}{0pt}
    \item Let $x$ be the question or context provided.
    \item Let $h$ be the image provided.
    \item Let $Y = \{y_1, y_2, \ldots, y_n\}$ be the set of multiple-choice options available.
    \item Let $y$ be the correct answer from $Y$.
\end{itemize}

The goal is to develop an in-context learning method that accurately selects $y$ from $Y$ given $x$ and $h$.

\subsection{Two-Step Scoring Method}

\subsubsection{Step 1: Elimination}

In the first step of the MM-PoE method, each option $y_i$ is scored based on a specified metric. The score function, $score(x, h, y_i)$, evaluates each option's plausibility given the question $x$ and image $h$. The scores are used to eliminate options that are deemed less likely to be correct. Specifically, options whose scores are below the average score are eliminated. This is calculated as follows:

$$
s_i = score(x, h, y_i)
$$
$$
Y_{\text{wrong}} = \{y_i | s_i < \text{avg}(s_1, \ldots, s_n)\}
$$

This elimination strategy intuitively aligns with how humans often discard options that seem clearly incorrect before carefully considering the remaining choices.

\subsubsection{Step 2: Prediction}

The second step involves making the final choice from the non-eliminated options. This step utilizes a binary mask to exclude the eliminated options during the prediction phase. The mask for each option $y_i$ is defined as follows:
$$
m_i = \begin{cases} 
0 & \text{if } y_i \in Y_{\text{wrong}} \\
1 & \text{otherwise}
\end{cases}
$$

The masked context $x_{\text{mask}}$ is then constructed by modifying the original context $x$ to include only the options for which $m_i = 1$. Each option is scored again, but this time within the context that explicitly excludes the eliminated options, possibly by using a template $T$ that masks out $Y_{\text{wrong}}$ in the presentation of the options:
$$
x_{\text{mask}} = T(x, Y, \text{mask})
$$

The final predicted answer $\hat{y}$ is then the option with the highest score among the remaining options:
$$
\hat{y} = \argmax_{i | m_i = 1} [score(x_{\text{mask}}, h, y_i)]
$$

\section{Experimental Setup}
\label{sec:experiments}
To evaluate the effectiveness of the MM-PoE, we designed an experimental framework that tests the method across a diverse set of visual reasoning datasets. This setup aims to compare MM-PoE with existing scoring methods to highlight its potential improvements in accuracy and reasoning capability. Following prior work on evaluating vision-language reasoning capabilities \cite{ma2023poe,lu2022learn}, we initially adopt a zero-shot experimental framework without any task-specific tuning to gauge the inherent reasoning strength of MM-PoE. Performance is measured using accuracy, averaged over multiple random seeds to ensure statistical robustness. Additionally, to examine the adaptability of MM-PoE to limited supervision scenarios, we also consider few-shot experiments where we aimed to observe any changes in effectiveness when provided with a small number of context-specific demonstrations are provided to the model.

\subsection{Benchmark Datasets}
\label{sec:data}
Our experiments were conducted on two different multiple-choice visual reasoning datasets - ScienceQA \cite{lu2022learn} and Diagram Understanding(AI2D) \cite{Kembhavi2016ADI}, selected to cover a broad spectrum of reasoning types and complexities. These tasks include both traditional visual reasoning tasks and more specialized ones designed to test specific reasoning skills. For example, AI2D focuses on diagrammatic reasoning, testing the model’s ability to interpret schematic information and select correct answers from multiple choices. To ensure a comprehensive evaluation, we used train sets from established benchmarks when available; otherwise, we utilized development sets. In the case of varying number of options in the multiple-choice answers for ScienceQA and AI2D datasets, we standardize the dataset by filtering questions containing image context and exactly four answer choices. The evaluation ensures that the comparisons reflect the intrinsic capabilities of MM-PoE without confounding effects from extensive fine-tuning.

\begin{table*}[!h]
\centering
\begin{tabular}{|c|c|c|c|c|}
\hline
\textbf{Dataset} & \textbf{\#Options} & \textbf{Train} & \textbf{Dev} & \textbf{Test} \\ \hline
ScienceQA        & 4                  & 12,726         & 4,241        & 4,241         \\ \hline
AI2D             & 4                  & 3,921          & 982          & -             \\ \hline
\end{tabular}
\caption{Summary of ScienceQA and AI2D datasets}
\label{tab:dataset_summary}
\end{table*}

\subsection{Models}
\label{sec:model}
For the core experiments, we utilized the GIT\cite{GIT_Base_VQA,GIT_Base_TextVQA} and BLIP\cite{BLIP2_FLAN,BLIP2_Opt,InstructBLIP} models, chosen for its balance between computational efficiency and performance in instruction-tuned vision language tasks. GIT, a generative image-to-text transformer, and BLIP, a bootstrapped language-image pre-training framework, represent state-of-the-art architectures capable of handling complex visual reasoning queries. These models have demonstrated strong capabilities in handling various multi-modal tasks and serves as a robust and reliable platform for evaluating our MM-PoE method.

\subsection{Baselines Approaches}

We compared MM-PoE against the following five baseline scoring methods to assess its relative performance:

\begin{itemize}\setlength{\itemsep}{0pt}
    \item \textbf{Language Modeling (LM)}: This baseline uses the raw vision language modeling likelihood as the scoring function.
    \item \textbf{Average Language Modeling (AVG)}: This method averages the log probabilities across all tokens in the option.
    \item \textbf{Calibration}: This involves adjusting the VLM scores based on calibration techniques that aim to correct for the model's confidence.
    \item \textbf{Channel}: Channel methods score each option based on how likely the question is given the option, which reverses the typical conditional probability used in VLMs.
    \item \textbf{Multiple Choice Prompting (MCP)}: This approach formats the input by presenting the question followed by all options, prompting the model to select the most likely option.
\end{itemize}

These baselines, as shown in Table \ref{tab:prompt_samples}, offer a comprehensive view of existing techniques, ranging from standard likelihood-based methods to more sophisticated calibration and prompting strategies. By positioning MM-PoE against these methodologies, we rigorously assess its relative strengths in structured elimination and final choice selection.

\subsection{Implementation}

The effectiveness of MM-PoE hinges on the robustness of the scoring function and the accuracy of the elimination step. The scoring function can be any VLM-based likelihood estimator, such as vision language modeling likelihood, or any of its alternatives like average log probability or calibrated log probability. Our implementation tests multiple such scoring functions to identify the most effective ones in both eliminating implausible options and accurately selecting the final answer. 

The MM-PoE method is designed to be model-agnostic, meaning it can be implemented using any existing VLM capable of scoring text options, and it is flexible enough to be adapted to different types of multiple-choice visual answering questions across various domains. The scoring functions were carefully chosen based on their theoretical alignment with the two-step elimination and prediction philosophy of PoE. We conducted extensive parameter tuning and optimization to maximize the performance of both the elimination step and the final prediction accuracy. All code\footnote{https://github.com/souradipp76/MM-PoE} and instructions for reproducing our experiments are publicly available, ensuring that interested researchers can integrate and refine MM-PoE within their own frameworks.

\section{Results}
\begin{figure*}[!h]
    \centering
    \begin{subfigure}[t]{0.5\textwidth}
        \centering
        \includegraphics[width=0.9\linewidth]{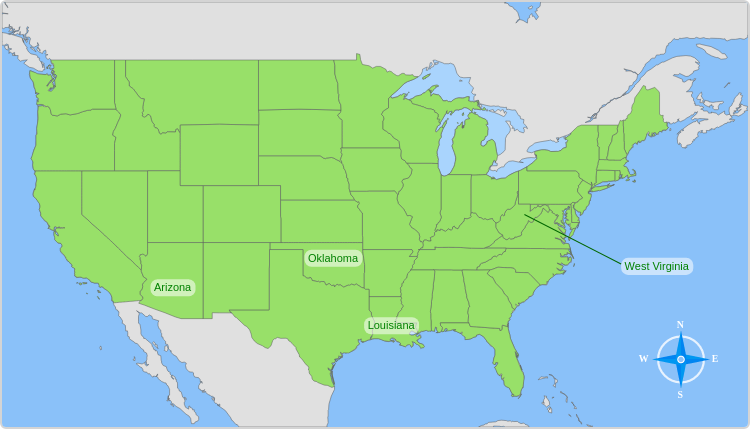}
        \caption{ScienceQA Example}
        \footnotesize{Question: Which of these states is farthest north?\\
        Options: West Virginia, Louisiana, Arizona, Oklahoma\\
        Ground Truth Option: West Virginia\\
        Predicted Masks: West Virginia, Louisiana, [MASK], [MASK]\\
        Predicted Option: West Virginia
        }
        \label{fig:scienceqa}
    \end{subfigure}%
    ~ 
    \begin{subfigure}[t]{0.5\textwidth}
        \centering
        \includegraphics[width=.55\linewidth]{17.png}
        \caption{AI2D Example}
        \footnotesize{Question: Are phytoplankton predators or prey in this food chain?\\
        Options: producer, predator, prey, NA\\
        Ground Truth Option: prey\\
        Predicted Masks: [MASK], predator, prey, NA\\
        Predicted Option: prey
        }
        \label{fig:ai2d}
    \end{subfigure}
    \caption{Examples showing the two-step scoring process for ScienceQA and AI2D dataset using GIT model}
    \label{fig:example}
\end{figure*}

Our experimental setup was designed to rigorously test the effectiveness of MM-PoE across a range of visual reasoning tasks and compare its performance against standard baseline methods. By thoroughly examining the interplay between scoring methods, model architectures, and the elimination strategy, our sample results, as shown in Figure \ref{fig:example}, demonstrate the generality and utility of MM-PoE in advancing multi-modal reasoning capabilities and the potential benefits of integrating a process of elimination approach into visual reasoning strategies for multiple-choice questions.

MM-PoE consistently outperformed or matched the best-performing baselines across all datasets, showing particular strength in logical reasoning. The method's effectiveness in separating elimination and prediction tasks was crucial to its success.

\begin{table*}[!h]
\centering
\begin{tabular}{|c|c|c|c|c|c|c|c|}
\hline
\textbf{Model} & \textbf{Dataset} & \textbf{LM} & \textbf{AVG} & \textbf{Calibration} & \textbf{Channel} & \textbf{MCP} & \textbf{MM-PoE} \\ \hline
    \multirow{2}{*}{microsoft/git-base-vqav2}  & ScienceQA & \textbf{27.4} & 17.8 & 23.2        & 24.6    & 25.8 & 27.2 \\\cline{2-8}
    & AI2D      & 25.4 & 26.2 & 26.4        & 25.4    & 25.3 & \textbf{26.5} \\ \hline
     \multirow{2}{*}{microsoft/git-base-textvqa} & ScienceQA & 21.8 & 20.4 & 25.8        & 23.4    & 23.6 & \textbf{28.2} \\\cline{2-8}
    & AI2D      & 26.5 & \textbf{27.6} & 20.8        & 26.2    & 24.2 & 26.8 \\ \hline
\end{tabular}
\caption{Comparison of GIT model accuracy across ScienceQA and AI2D datasets for MM-PoE with respect to the baseline scoring methods.}
\label{tab:model_performance}
\end{table*}

\begin{table*}[!h]
\centering
\begin{tabular}{|c|c|c|c|c|}
\hline
\textbf{Model} & \textbf{Dataset} & \textbf{n\_shot} & \textbf{Mask Accuracy} & \textbf{MM-PoE} \\ \hline
    \multirow{6}{*}{microsoft/git-base-vqav2}  & \multirow{3}{*}{ScienceQA} & 0 & 75.9 & \textbf{23.8} \\ \cline{3-5}
    & & 1 & \textbf{76.8} & 22.4 \\ \cline{3-5}
    & & 3 & 76 & 21.4 \\ \cline{2-5}
    & \multirow{3}{*}{AI2D} & 0 & \textbf{74.1} & 24.3 \\ \cline{3-5}
    & & 1 & 71 & \textbf{24.4}\\ \cline{3-5}
    & & 3 & 71 & 24.2\\\hline
    \multirow{6}{*}{microsoft/git-base-textvqa}  & \multirow{3}{*}{ScienceQA} & 0 & \textbf{79.6} & 24.4 \\ \cline{3-5}
    & & 1 & 78.6 & 23.7 \\ \cline{3-5}
    & & 3 & 75 & \textbf{25} \\ \cline{2-5}
    & \multirow{3}{*}{AI2D} & 0 & 71 & 24 \\ \cline{3-5}
    & & 1 & \textbf{75} & \textbf{25.8}\\ \cline{3-5}
    & & 3 & 72.5 & 24\\\hline
\end{tabular}
\caption{Comparison of GIT model accuracy across ScienceQA and AI2D datasets for MM-PoE with zero-shot and n-shot($n=(1,3)$) prompting along with the masking accuracy after the first step.}
\label{tab:fewshot_performance}
\end{table*}

\section{Conclusion}
\label{sec:conclusion}

This paper introduces the Multi-Modal Process of Elimination (MM-PoE), an innovative approach designed to enhance the reasoning capabilities of Vision-Language Models (VLMs) in multiple-choice visual reasoning tasks. By adopting human-like test-taking strategies—specifically, the elimination of patently incorrect options prior to refining the final choice — MM-PoE achieves notable improvements in accuracy and robustness as seen in Table \ref{tab:model_performance}. Even in the case of few-shot experiments, our approach shows comparable performance with respect to the zero-shot case. It is to be noted that a better masking accuracy after Step 1 of the two-step scoring process, may not necessarily lead to better final score the models cannot completely ignore the masked options provided as inputs in Step 2. In general, a higher masking accuracy can surely convince us that the model is practically eliminating the incorrect options as expected. Thus, this method effectively addresses two critical shortcomings of the previous research \cite{ma2023poe} by not only extending the investigation to few-shot settings from zero-shot scenarios but also by introducing a multi-modal dimension to the previously single-modality, language-only framework. Future research directions include refining the elimination criteria, removing masked options, and investigating more sophisticated scoring functions. MM-PoE can also potentially be integrated into broader multi-modal pipelines and extended to advanced domains such as medical imaging and complex scientific classification, where robust multiple-choice reasoning is critically important.

\bibliographystyle{unsrt} 
\bibliography{references}

\appendix

\section{Ethical Limitations}
While MM-PoE offers a step towards more interpretable and accurate reasoning in multi-modal AI systems, potential ethical concerns remain. Over-reliance on model-generated reasoning may lead to unintentional biases or misinterpretations, particularly if datasets contain skewed examples or sensitive visual content. Additionally, as MM-PoE and related techniques become more capable, users must remain vigilant about ensuring that these models are applied in ethically responsible ways. Future work will focus on developing safeguards, including fairness audits and transparency tools, to mitigate potential harms and ensure that improvements in reasoning quality do not come at the expense of ethical integrity.

\section{Scoring Methods \& Prompt Samples}

\begin{table*}[!h]
  \centering
  \begin{tabular}{|c|c|c|c|}
  \hline
    \textbf{Method}     & \textbf{Score}     & \textbf{Input}  & \textbf{Output} \\\hline
    LM & $\log P (y_i|x,h)$ & $x, \langle h \rangle$ the answer is: & $y_1$ \\\hline
    AVG & $1/len(y_i)*\log P(y_i|x,h)$ & $x, \langle h \rangle$ the answer is: & $y_1$ \\\hline
    Calibration & $\log P(y_i|x,h)$ & $x, \langle h \rangle$ the answer is: & $y_1$ \\\hline
    Channel & $\log P(x|y_i,h)$ & $y_1, \langle h \rangle$ & $x$ the answer is: \\\hline
    \multirow{5}{*}{MCP} & \multirow{5}{*}{$\log P (y_i|x,h, Y )$} &
    Question: $x$, Image: $\langle h \rangle$ & \multirow{5}{*}{A} \\
    & & A. $y_1$ & \\
    & & B. $y_2$ & \\
    & & C. $y_3$ & \\
    & & Answer: & \\\hline
    \multirow{5}{*}{MM-POE: Elimination} & \multirow{5}{*}{$\log P (y_i|x,h, Y )$} &
    Question: $x$, Image: $\langle h \rangle$ & \multirow{5}{*}{A} \\
    & & A. $y_1$ & \\
    & & B. $y_2$ & \\
    & & C. $y_3$ & \\
    & & Answer: & \\\hline
    \multirow{8}{*}{MM-POE: Prediction} & \multirow{8}{*}{$\log P (y_i|x_{mask},h, Y \setminus Y_{wrong})$} &
    Select the most suitable option & \multirow{8}{*}{B} \\
    & & to answer the question. &\\
    & & Ignore [MASK]  options. &\\
    & & Question: $x$, Image: $\langle h \rangle$ & \\
    & & A. [MASK] & \\
    & & B. $y_2$ & \\
    & & C. $y_3$ & \\
    & & Answer: & \\\hline
  \end{tabular}
  \caption{Input and output samples for the different  scoring methods}
  \label{tab:prompt_samples}
\end{table*}

\end{document}